\documentclass[screen]{acmart}

\author{Nurit Cohen-Inger}
\orcid{0009-0009-3227-3090}
\email{cohening@post.bgu.ac.il}
\author{Seffi Cohen}
\orcid{0000-0002-1135-0079}
\email{seffi@post.bgu.ac.il}
\author{Neomi Rabaev}
\email{rabaevn@post.bgu.ac.il}
\author{Lior Rokach}
\orcid{0000-0002-6956-3341}
\email{liorrk@post.bgu.ac.il}
\author{Bracha Shapira}
\orcid{0000-0003-4943-9324}
\email{bshapira@bgu.ac.il}

\affiliation{%
  \institution{Ben Gurion University}
  \city{Beer Sheva}
  \country{Israel}
  \postcode{8410501}
}

\setcopyright{acmlicensed}

\DeclareUnicodeCharacter{05BF}{} 
\usepackage{booktabs}
\usepackage{multirow}
\usepackage{amsmath}
\usepackage{algorithm}
\usepackage{algorithmic}
\usepackage{newfloat}
\usepackage{listings}
\usepackage{natbib}
\usepackage{graphicx}
\usepackage{url}

\begin{document}

\title{BiasGuard: Guardrailing Fairness in Machine Learning Production Systems}

\begin{abstract}
As machine learning (ML) systems increasingly impact critical sectors such as hiring, financial risk assessments, and criminal justice, the imperative to ensure fairness has intensified due to potential negative implications. While much ML fairness research has focused on enhancing training data and processes, addressing the outputs of already deployed systems has received less attention. This paper introduces 'BiasGuard', a novel approach designed to act as a fairness guardrail in production ML systems. BiasGuard leverages Test-Time Augmentation (TTA) powered by Conditional Generative Adversarial Network (CTGAN), a cutting-edge generative AI model, to synthesize data samples conditioned on inverted protected attribute values, thereby promoting equitable outcomes across diverse groups. This method aims to provide equal opportunities for both privileged and unprivileged groups while significantly enhancing the fairness metrics of deployed systems without the need for retraining. Our comprehensive experimental analysis across diverse datasets reveals that BiasGuard enhances fairness by 31\% while only reducing accuracy by 0.09\% compared to non-mitigated benchmarks. Additionally, BiasGuard outperforms existing post-processing methods in improving fairness, positioning it as an effective tool to safeguard against biases when retraining the model is impractical.
\end{abstract}


\maketitle


\section{Introduction}
\label{sec:intro}
In the evolving landscape of machine learning applications, the imperative to ensure fairness has intensified, particularly given their substantial impact on society \cite{caton2020fairness}. The pursuit of fairness aims to achieve equitable outcomes for all individuals and groups, a goal often compromised by inherent biases that manifest themselves as disparate impacts based on protected attributes such as race or gender \cite{holstein2019improving}. These biases risk perpetuating discrimination in essential areas including employment, legal sentencing, and educational admissions \cite{mehrabi2021survey}, highlighting the urgent need for effective interventions to address these disparities and promote fairness \cite{barocas2023fairness}. Protected attributes are characteristics that are legally or socially protected from discrimination, such as race, gender, age, religion, sexual orientation, disability status, and national origin. Therefore, it is essential that machine learning models make decisions that are not only transparent but also consistently fair, ensuring that both privileged and marginalized groups are safeguarded from biased outcomes.

Traditional bias mitigation strategies include pre-processing, in-processing, and post-processing techniques \cite{hort2022bia}. However, in many real-world applications, especially those involving pre-trained Software as a Service (SaaS) models or closed-source models \cite{lewicki2023out}, modifying the training data or adjusting the training procedures is often impractical or impossible. Moreover, biases can arise due to changes in data distributions or changes in social norms within production environments, which are not represented in the training set \cite{torpmann2024robust}. These challenges necessitate a robust post-processing approach that can ensure fairness in production environments without requiring alterations to the foundational components of the model.

This paper introduces 'BiasGuard', an innovative post-processing method that employs Test-Time Augmentation to generate augmented test samples, facilitating fairer predictions in production environments. 

\textbf{Motivating BiasGuard:} BiasGuard is distinguished by its use of TTA to create augmented test samples that dynamically balance fairness. By leveraging the CTGAN to generate synthetic data conditioned on the inverse values of protected attributes, By integrating synthetic data generated through CTGAN, a generative AI model, BiasGuard introduces a novel mechanism for dynamically recalibrating predictions near decision boundaries, enhancing fairness in uncertain scenarios This method ensures that both privileged and unprivileged groups receive fair treatment by adapting predictions in real-time, thereby providing equal opportunities across all demographics. To our knowledge, this is the first implementation of a bias mitigation strategy that enhances test data post-deployment through synthetic data, addressing fairness within the constraints of operational ML systems.

\begin{figure}[tbh!]
    \centering
    \includegraphics[scale=0.9]{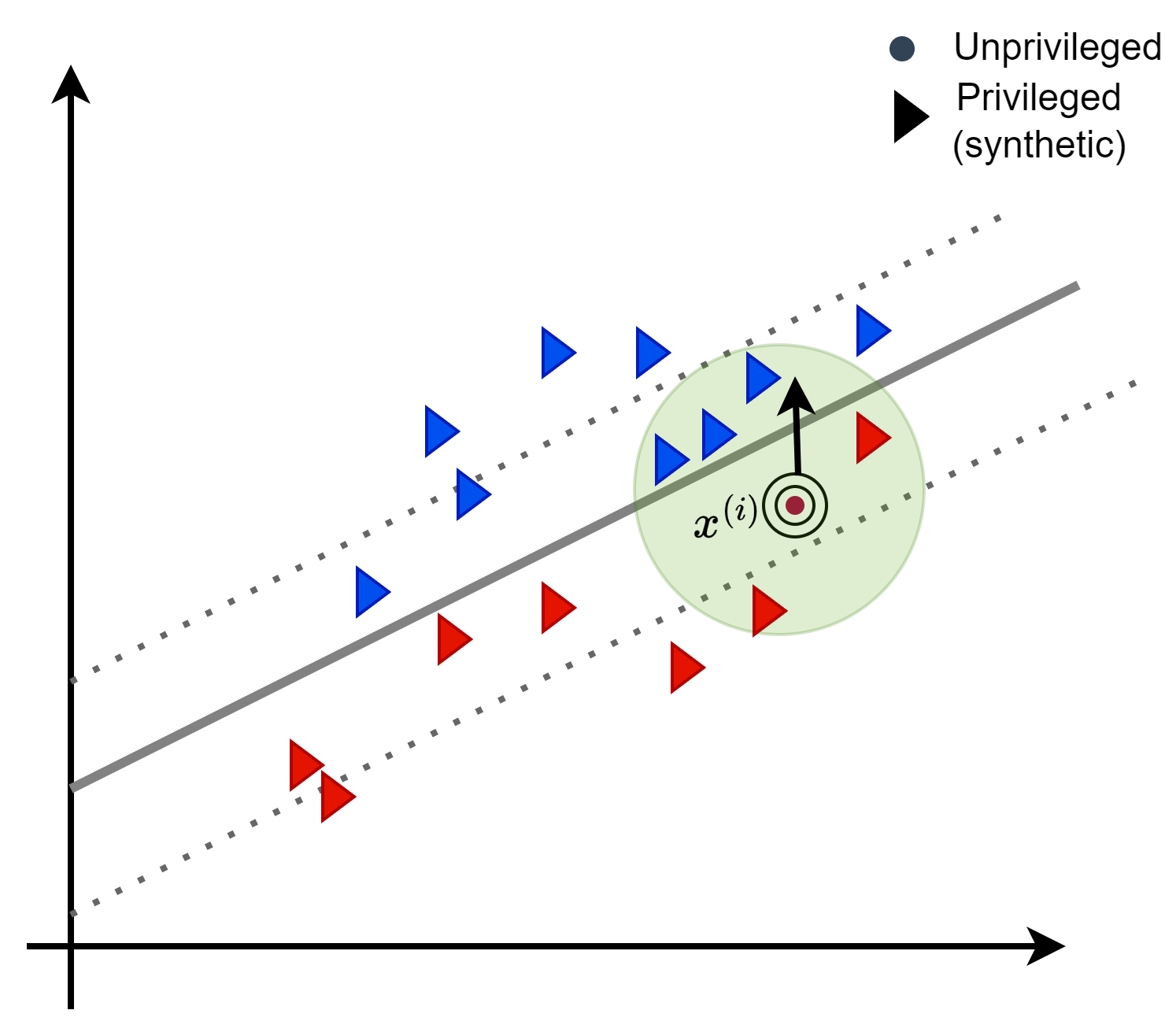}
    \caption{BiasGuard motivation - For every sample \(x^{(i)}\) from the test set, synthetic data is generated with the opposite value of its protected attribute as a condition. Then, the prediction is balanced with the nearest samples.}
    \label{fig:motivation}
\end{figure}

\textbf{Our key contributions are:}
\begin{itemize}
\item \textbf{An effective post-processing bias mitigation method:} We demonstrate how BiasGuard improves fairness metrics in a variety of datasets. This method is especially crucial in production environments where retraining the model is impossible and data may deviate from the training set conditions. BiasGuard serves as a robust fairness guardrail, ensuring equitable outcomes without requiring modifications to the underlying model training.

\item \textbf{Model-agnostic design of BiasGuard:} BiasGuard operates independently of the underlying model specifics, ensuring its adaptability across various ML frameworks while preserving the integrity of the model as a black box. This versatility allows for integration without needing access to or modifications of the model's training architecture, making it ideal for maintaining fairness in dynamic production settings.

\item \textbf{Reducing bias by utilizing generative AI technology:} We use synthetic data generated by CTGAN to significantly reduce bias, replicating predicted samples with the opposite protected attribute and making adjustments to ensure equitable opportunities. This method effectively addresses fairness challenges, adapting to evolving operational conditions and data characteristics in real time.

\end{itemize}

\section{Background and Related Work}
\label{sec:background}
This section covers the background and relevant work to our research, focusing on detecting and mitigating bias in ML systems, with particular emphasis on post-processing mitigation methods that are relevant for production scenarios. Additionally, since our method employs Test-Time-Augmentation using the CTGAN technique to create synthetic test data, we will review this background as well.

\subsection{Bias Detection Metrics}

Canton et al.'s survey \cite{caton2020fairness} identifies two primary categories of bias metrics: those based on outcome probabilities, such as Statistical Parity \cite{calders2010three} and Disparate Impact \cite{feldman2015certifying}, and those derived from the confusion matrix, such as Equalized Odds and Opportunity \cite{hardt2016equality}, and Accuracy Rate Difference \cite{berk2021fairness}. A single scenario can be evaluated differently on various metrics, some considered fair and others unfair \cite{verma2018fairness}.

Our selection of the Equalized Odds metric, as the leading metric, was guided by its ability to comprehensively evaluate correct and incorrect classifications between groups, ensuring comparable True Positive Rates (TPR) and False Positive Rates (FPR). This approach provides a more nuanced examination of fairness compared to metrics that focus on a single type of error. It avoids the pitfalls of statistical metrics like Disparate Impact, which only considers outcome proportions and could mislead by promoting uniform denials \cite{hardt2016equality}. Additionally, it steers clear of the over correction risks associated with Disparate Impact, which might favor underrepresented groups without genuinely addressing fairness \cite{corbett2023measure}. Confusion-matrix based metrics, especially suitable for post-processing, do not require manipulation of the training dataset and account for 'real-life' differences between groups. Within this category, Equalized Odds offers a broader perspective by addressing both false positives and false negatives, directly tackling the key issues of biases that may occur.

\subsection{Bias Mitigation Methods}

The field of bias mitigation has seen significant research growth, with approximately 350 related studies \cite{hort2022bia} categorized into pre-processing, in-processing, and post-processing approaches:

\textbf{Pre-processing methods} aim to eliminate bias from the training data by transforming the protected attribute or balancing groups \cite{feldman2015certifying, calmon2017optimized, lum2016statistical}, learning fair representation \cite{samadi2018price}, or changing protected attribute distribution \cite{friedler2014certifying}. FairUS method \cite{coheninger2024fairus} recently used CTGAN technique to upsample the training set.

\textbf{In-processing methods} incorporate fairness constraints during training. A relevant technique uses GAN technology to debias the model's outcome \cite{abusitta2019generative}. However, our approach differs as we employ GANs to generate synthetic data for TTA, not during model training.

\textbf{Post-processing approaches}, our primary focus, apply transformations to the model output to enhance prediction fairness. These methods are ideal for settings where the ML pipeline is a black box and direct model adjustments are unfeasible. A recent study highlights that only a small fraction (approximately 16\%) of the research has explored post-processing techniques for bias mitigation \cite{hort2022bia}, indicating this area for innovation. These strategies range widely, including decision reversal \cite{heidari2018preventing}, threshold adjustments for demographic parity \cite{menon2018cost, corbett2017algorithmic, kobayashi2021one}, and enhancements for equal opportunities and odds \cite{hardt2016equality, ying2024improving}.

BiasGuard differentiates itself by focusing on post-processing fairness in production environments. Unlike other post-processing methods that typically adjust model outputs directly, BiasGuard uniquely enhances the model's predictions by integrating synthetic data produced by generative AI techniques, offering a novel and effective means to recalibrate fairness dynamically in production systems.

\subsection{Test-Time Augmentation}

Test-time augmentation is an enhancement method applied during the testing phase instead of training. This process, which generates multiple augmented versions of each test sample and then amalgamates these outcomes with those of the unaltered sample, is particularly beneficial in scenarios where model retraining or modification is impractical. The adoption of TTA has been notable in fields such as computer vision, natural language processing \cite{cohen2023enhancing, cohen2024improving}, and anomaly detection \cite{cohen2023boosting, cohen2023ttanad}, demonstrating its versatility and effectiveness in improving model performance without structural changes.

\subsubsection{GAN-based Augmentation for Tabular Data}

The Conditional Generative Adversarial Network (CTGAN) introduced by Xu et al. \cite{xu2019modeling} represents a significant advancement in generating synthetic tabular data, adept at mimicking complex distributions. CTGAN, as an advanced generative AI technique, plays a pivotal role in BiasGuard, synthesizing data that preserves original feature distributions while conditioning on inverse protected attributes to address fairness, thus facilitating a nuanced adjustment of biases at test time, ensuring the model remains a black box while evaluated under fairer conditions.

\section{Method}
\label{sec:method}
\begin{figure*}[tbh!]
    \centering
    \includegraphics[width=\textwidth]{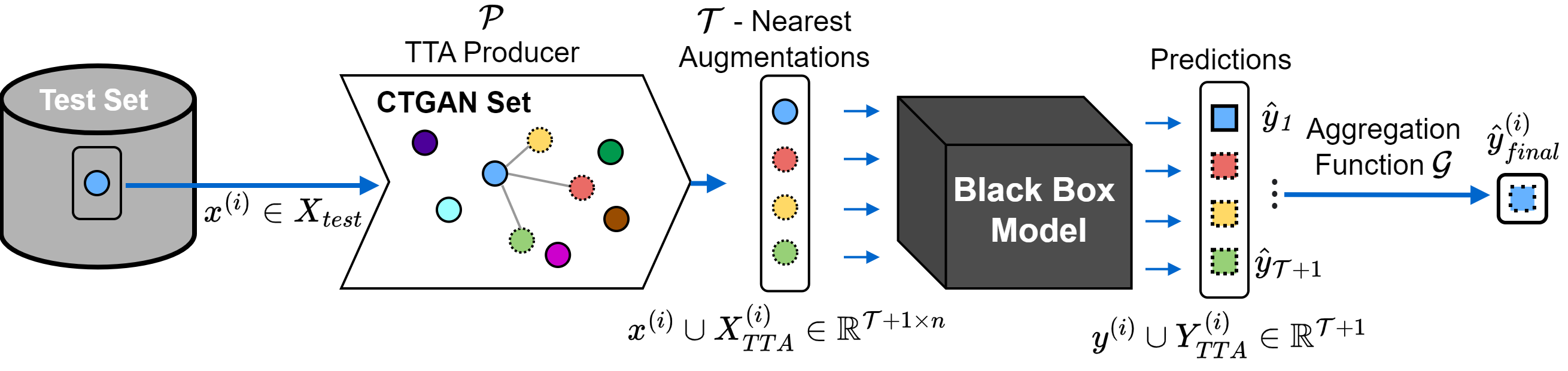}
    \caption{An overview of the BiasGuard method. \textbf{1} - For every sample \(x^{(i)}\) from the test set. \textbf{2} - A TTA of synthetic data based on CTGAN is chosen with the opposite protected value of \(x^{(i)}\). \textbf{3} - TTA predictions are received from the black-box model. \textbf{4} - Generation of the final prediction \(\hat{y}_{final}\) by aggregating the instance prediction with all the augmentation predictions.}
    \label{fig:phase1phase2overview}
\end{figure*}

BiasGuard provides a systematic approach to fair classification in production, by leveraging augmented test (inference) samples with opposite protected attributes (e.g., "male" replaced by "female"). This method averages the probabilities of several samples, regardless of the protected value (privileged and unprivileged), to achieve fairer predictions. BiasGuard is designed to actively monitor and adjust the fairness of predictions in production, ensuring ongoing compliance with fairness standards. Figure \ref{fig:phase1phase2overview} illustrates the BiasGuard method applied during the test phase. For each instance in the test set, BiasGuard generates a balanced subset of augmentations with the inverse protected attribute value, then classifies both the augmentations and the original instance using a black-box model.

\subsection{Formulation}

Consider a test set $X_{test} \in \mathbb{R}^{m \times n}$, where $m$ denotes the number of instances and $n$ the number of attributes. Let $y \in [0,1]^m$ represent the label vector for $X_{test}$. We define $\mathcal{M}: \mathbb{R}^{m \times n} \rightarrow [0,1]^m$ as a pre-trained black-box binary classifier that generates classification probabilities $\hat{y} \in [0,1]^m$ for the test set $X_{test}$.

Our focus is on enhancing fairness in binary classification models across a range of protected attributes. For instance, in tasks such as predicting criminal recidivism, let $y \in Y = \{0,1\}$ be the target label, and $PA \in  \{0,1\}$ the protected attribute, where $y = 1$ denotes a favorable outcome (e.g., lower risk of recidivism), $PA = 1$ a privileged group (e.g., Caucasian for the race protected attribute), and $PA=0$ an unprivileged group (e.g., non-Caucasian). Our method is demonstrated on binary protected attributes and can be extended to multi-class classification and any number of categories for the protected attribute using one-hot encoding.

The process of TTA involves generating test-time augmentations for each test instance $x^{(i)} \in X_{test}$ where $0 \leq i \leq m$, denoted by $X^{(i)}_{TTA} \in \mathbb{R}^{\mathcal{T} \times n}$, where $\mathcal{T}$ is the number of augmentations produced by the CTGAN. We denote $\hat{y}^{(i)} = \mathcal{M}(x^{(i)}) \in [0,1]$ and $\hat{Y}^{(i)}_{TTA} = \mathcal{M}(X^{(i)}_{TTA}) \in [0,1]^{\mathcal{T}}$ as the predictions of $x^{(i)}$ and $X^{(i)}_{TTA}$, respectively, using the classifier $\mathcal{M}$. An aggregation function:
\begin{equation}
    \mathcal{G} : ([0,1], [0,1]^{\mathcal{T}}) \rightarrow [0,1]
\end{equation}
(e.g., weighted average) combines the original's and TTA's predictions to derive the final prediction for test instance \(x^{(i)}\):
\begin{equation}
    \hat{y}_{final}^{(i)} = \mathcal{G}(\hat{y}^{(i)},\hat{Y}^{(i)}_{TTA}) \in [0,1].
\end{equation}

\begin{algorithm}[h]

\caption{BiasGuard}
\label{alg:BiasGuard_algorithm}
\begin{algorithmic}[1]
\STATE \textbf{Input}: 
    \begin{itemize}
        \item \(PA\): Protected Attribute with (\(\text{Privileged}\) ,\(\text{Unprivileged}\))
        \item \(\mathcal{T}\): Number of augmentations to generate
        \item  \(\mathcal{G}\): Aggregation function for balanced prediction with augmentations 
        \item \(CTGAN_{\text{PA}}\): TTA producers (one for each Protected attribute value)  
        \item \(X_{test}\): Test set
        \item \(\mathcal{M}\): Black-box classifier
        
    \end{itemize}

\STATE \textbf{Output}: 
    \begin{itemize}
        \item \(\hat{Y}_{final}\): Fair predictions array for all instances in \(X_{test}\)
    \end{itemize}

\STATE Initialize \(\hat{y}_{final}, \hat{Y}_{final}, \hat{Y}^{(i)}_{\text{TTA}} \gets \emptyset\)

\FOR{each test instance \(x^{(i)} \in X_{test}\)}
    
    \STATE // 1. Create an "opposite" version of the protected attribute for \(x^{(i)}\)

    \STATE \( x_{\text{opposite}}^{(i)} \gets x^{(i)}_{\text{\( \neg PA\)}} \)

    \STATE // 2. Get the classifier predictions for both versions of the instance
    \STATE \(\hat{y}^{(i)} = \mathcal{M}\bigl(x^{(i)}\bigr)\)
    \STATE \(\hat{y}^{(i)}_{\text{opposite}} = \mathcal{M}\bigl(x_{\text{opposite}}^{(i)}\bigr)\)

    \STATE // 3. Check if there is a potential for bias
\IF{\(\mathrm{round}\bigl(\hat{y}^{(i)}\bigr) \neq \mathrm{round}\bigl(\hat{y}^{(i)}_{\text{opposite}}\bigr)\)}
        \STATE \(X^{(i)}_{\text{TTA}} \gets \mathrm{NearestNeighbors}\Bigl(x^{(i)}, CTGAN_{\text{PA}}, \mathcal{T}\Bigr)\)
        \STATE \(\hat{Y}^{(i)}_{\text{TTA}} \gets \mathcal{M}\bigl(X^{(i)}_{\text{TTA}}\bigr)\)
    \ENDIF
    \STATE \(\hat{y}^{(i)}_{\text{final}} \gets \mathcal{G}\Bigl(\hat{y}^{(i)}, \hat{Y}^{(i)}_{\text{TTA}}\Bigr)\)

    \STATE // 4. Append the final balanced prediction to the array of predictions
    \STATE \(\hat{Y}_{final} \gets \hat{Y}_{final} \cup \{\hat{y}^{(i)}_{\text{final}}\}\)

\ENDFOR

\STATE \textbf{return} \(\hat{Y}_{final}\)
\end{algorithmic}
\end{algorithm}

\textbf{Algorithm \ref{alg:BiasGuard_algorithm} Explained:}
BiasGuard is a model-agnostic bias mitigation method that operates purely on the outputs of a black-box classifier~\(\mathcal{M}\). The Algorithm inputs are:
\begin{enumerate}
    \item A protected attribute \(PA\) with values \(\text{Privileged}\) and \(\text{Unprivileged}\).
    \item A test set \(X_{test}\).
    \item A number of test-time augmentations \(\mathcal{T}\).
    \item A set of CTGAN generators, \(CTGAN_{\text{PA}}\), one per value of \(PA\).
    \item An aggregation function \(\mathcal{G}\) for combining the predictions.
\end{enumerate}

For each instance \(x^{(i)}\in X_{test}\), the algorithm first constructs an opposite version \(x_{\text{opposite}}^{(i)}\) by flipping only the protected attribute value (line~6). The predictions \(\hat{y}^{(i)}\) and \(\hat{y}^{(i)}_{\text{opposite}}\) are obtained by feeding each version of the instance to the black-box classifier~\(\mathcal{M}\) (lines~8--9).

Next, BiasGuard checks if the predicated classes differ (line~10) by rounding each predicted probability. If \(\mathrm{round}\bigl(\hat{y}^{(i)}\bigr)\neq \mathrm{round}\bigl(\hat{y}^{(i)}_{\text{opposite}}\bigr)\), it concludes there is a potential for bias derived from the \(PA\). In that case, the method proceeds to generate \(\mathcal{T}\) synthetic samples with the opposite value of the protected attribute using a CTGAN (line~12) and then obtains predictions on these augmentations (line~13). If there is no difference in the discrete labels, the algorithm uses only the original prediction without extra augmentations.

Finally, BiasGuard applies its aggregation function~\(\mathcal{G}\) (line~15) to combine the original prediction~\(\hat{y}^{(i)}\) with the average of the TTA predictions~\(\hat{Y}^{(i)}_{\text{TTA}}\). By default, we define \(\mathcal{G}\) as:
\[
    \mathcal{G} \bigl(\hat{y}^{(i)}, \hat{Y}^{(i)}_{\text{TTA}}\bigr) \;=\; 
    \frac{1}{2} \Bigl(
        \hat{y}^{(i)} 
        \;+\; 
        \frac{\sum \hat{Y}^{(i)}_{\text{TTA}}}{|\mathcal{T}|}
    \Bigr),
\]
though other forms of aggregation (e.g., majority vote) could be used. The final, balanced prediction \(\hat{y}^{(i)}_{\text{final}}\) is then appended to \(\hat{Y}_{final}\). Once the loop completes, \(\hat{Y}_{final}\) comprises the fair predictions for the entire test set.


\subsection{Why BiasGuard Improves Fairness}

Consider two nearly identical individuals who differ only by their protected attribute (Privileged vs.\ Unprivileged). If the predicted outcome changes from favorable (e.g., “selected”) for the Privileged individual to unfavorable (e.g., “not selected”) for the Unprivileged one, this suggests potential bias because the difference in the decision is driven solely by the protected attribute rather than relevant competencies or other features.

BiasGuard detects this situation in inference time by comparing predictions for the original instance \(x^{(i)}\) and its opposite-protected-attribute version \(x_{\text{opposite}}^{(i)}\). If the label flips when the protected attribute is switched, BiasGuard infers there may be bias and proceeds to generate additional TTA from the opposite group. It then balances (aggregates) the original and augmented predictions to “smooth out” any unjustified discrepancy. Consequently, the final decision is more robust to the influence of the protected attribute, reducing bias without changing the underlying model.

From a fairness metric perspective, BiasGuard lowers the gap between the true positive rate (TPR) and false positive rate (FPR) for Privileged vs.\ Unprivileged groups. Equalized Odds (EOD) measures these rate differences, so when BiasGuard raises TPR where it was too low or lowers FPR where it was too high (for individuals triggered by the opposite-attribute check), it naturally decreases the disparity between Privileged and Unprivileged outcomes. As these disparities shrink, EOD moves closer to zero, indicating a fairer model under this definition of fairness.

\subsection{Time Complexity}
The computational efficiency of BiasGuard is a critical factor for real-world
deployment, particularly in production settings with strict latency requirements.
BiasGuard’s complexity arises mainly from synthetic data generation and subsequent
prediction on augmented samples. First, for each test instance, BiasGuard uses the
CTGAN generator (with \(L\) layers and input dimensionality \(D\)) to produce
\(T\) synthetic augmentations; each synthetic instance generation costs
\(O(L \cdot D^2)\). Thus, for \(m\) instances, total synthetic data generation
takes \(O(mT \cdot L \cdot D^2)\). Although this overhead surpasses that of
simpler baselines, it can be mitigated using GPUs or lightweight generative models.
Next, predictions for the original and augmented samples require \(O(mT)\) time
since each instance and its synthetic variants are fed into the black-box model,
and their outputs are aggregated (e.g., via weighted averaging). Sensitivity
analyses show that reducing \(T\) decreases latency while still achieving
substantial fairness gains, making BiasGuard amenable to large-scale, real-time
applications.

\section{Experiments}
\label{sec:experiments}
The experiments section evaluates the BiasGuard method's performance in minimizing Equalized Odds while maintaining a satisfactory level of accuracy. This section describes the experimental setup, including methods, classifier, configurations, and datasets used.
We conducted five experiments to assess BiasGuard's ability to minimize Equalized Odds with only a minor effect on the model's performance (accuracy). Each experiment yields 7 results, which include the measurements of Equalized Odds and accuracy for three compared baselines and our method in several configurations.

\subsection{Evaluation Metrics}
For our method evaluation, we utilize two primary categories of fairness metrics: confusion-matrix-based measures and outcome-based measures, along with accuracy as follows:

\begin{itemize}
    \item \textbf{Equalized Odds (EOD).}
    As discussed in the background section, EOD \cite{hardt2016equality} belongs to the confusion-matrix category, capturing discrepancies in both false positive and true positive rates. Let \(PA\) be the protected attribute with values (\(\text{Privileged}\), \(\text{Unprivileged}\)). Then:
    \begin{equation}
    EOD = \frac{1}{2}
    \Bigl(
    \bigl|\mathrm{TPR}_{PA=\text{Privileged}} - \mathrm{TPR}_{PA=\text{Unprivileged}}\bigr|
    + 
    \bigl|\mathrm{FPR}_{PA=\text{Privileged}} - \mathrm{FPR}_{PA=\text{Unprivileged}}\bigr|
    \Bigr).
    \end{equation}
    By emphasizing both TPR and FPR, EOD offers a nuanced view of bias that goes beyond assessing only one type of error.

    \item \textbf{Disparate Impact ($DI$).}
    In contrast, Disparate Impact falls under the outcome-based (or ``statistical'') category, as discussed in the background section \cite{caton2020fairness}. It measures the ratio of positive decisions received by unprivileged and privileged groups:
    \begin{equation}
    DI = \frac{\Pr(\hat{Y} = 1 \mid PA=\text{Unprivileged})}
              {\Pr(\hat{Y} = 1 \mid PA=\text{Privileged})}.
    \end{equation}
    A value of \(DI\) close to 1 indicates similar treatment across the two groups, whereas values significantly different from 1 may indicate disparate treatment.

    \item \textbf{Accuracy.}
    Since bias mitigation can sometimes reduce predictive performance, we also track the model’s overall accuracy. This provides insight into how fairness interventions interact with predictive effectiveness.

\end{itemize}

To ensure the stability of our results, we compute the standard deviation (STD) for each metric. By analyzing EOD, DI, and accuracy side by side, we capture both the effectiveness of our BiasGuard method in mitigating bias and its impact on model performance. 

\subsection{Compared Methods}

The experiments are performed on five benchmark datasets, using the proposed BiasGuard method with different number of augmentations (named BiasGuard-2, BiasGuard-4, BiasGuard-6 and BiasGuard-8) against three compared methods, namely:

\begin{enumerate}

    \item Baseline - The original model without any mitigation technique as a baseline.
    \item Reject Option - The original model with the Reject Option post-processing method. 
    \item Threshold Opt - The original model with Threshold Optimizer post-processing method.

\end{enumerate} 

We selected the Reject Option and the Threshold Optimizer methods for our baseline comparison due to their well-established foundation in fairness literature and proven state-of-the-art performance in recent studies, such as \cite{sikdar2022getfair}. Their suitability for black-box models and alignment with our focus on post-processing methods that change the predictions where uncertainty exists, make them ideal benchmarks to assess the incremental benefits of our BiasGuard method in enhancing fairness without compromising accuracy.

\subsection{Implementation Details}
To demonstrate BiasGuard we employed a Random Forest model from Sklearn \cite{pedregosa2011scikit} library, which aimed to evaluate the impact of the mitigation on fairness and accuracy metrics. Hort et al. \cite{hort2022bia} show that Random Forest is the common classifier used in bias mitigation studies. It should be noted that our method is model agnostic and we used the model for empirical proof of our concept.

We used the "Synthetic Data Vault Project" (SDV) CTGAN implementation\footnote{\url{https://github.com/sdv-dev/CTGAN}} with the following tuned hyper-parameters: 500 epochs, embedding\_dim set to 32, generator\_dim and discriminator\_dim set to (256,128,64,32), discriminator and generator learning rate set to 5e-6.

\subsection{Datasets}
We selected a variety of datasets \cite{le2022survey} from multiple domains (legal, medical, recruitment, and census) to showcase BiasGuard’s adaptability and efficacy across different protected attributes. Each dataset poses distinct fairness challenges, involving attributes such as \emph{sex}, \emph{age}, and \emph{race}, illustrating the broad applicability of BiasGuard. By evaluating our method on such diverse and domain-specific tasks, we aim to validate its robustness and effectiveness in real-world scenarios.

\textbf{LAW (SEX).}
This dataset contains \textbf{20,798} samples with 12 attributes, where \emph{sex} is the protected attribute \cite{wightman1998lsac}. The prediction task typically relates to assessing admission outcomes, making it a critical domain where equity is essential.

\textbf{SURGICAL (AGE).}
Related to a medical contextֿ\footnote{\url {https://www.kaggle.com/datasets/omnamahshivai/surgical-dataset-binary-classification/data}}, this dataset contains \textbf{14,635} samples with 25 attributes. Here, the protected attribute is \emph{age}, and the target variable involves predicting surgical outcomes. Fairness considerations are vital to ensure equitable treatment recommendations for different age groups.

\textbf{Utrecht RECRUIT (SEX).}
Originating from a synthetic recruitment scenario\footnote{\url{https://www.kaggle.com/datasets/ictinstitute/utrecht-fairness-recruitment-dataset}}, this dataset includes \textbf{4,000} samples with 15 attributes. Where \emph{gender} is the protected attribute, predicting whether a candidate will be hired. This domain demonstrates how algorithmic decisions can impact employment aspects.

\textbf{ADULT (RACE).}
Commonly referred to as the census income dataset,\footnote{\url{http://archive.ics.uci.edu/ml}} it comprises \textbf{32,561} post-cleaning samples \cite{kohavi1996scaling} with 15 attributes. The task is to predict whether an individual’s income exceeds \$50K per year. We focus on mitigating bias concerning \emph{race} as the protected attribute.

\textbf{COMPAS (SEX).}
A well-known dataset for predicting criminal recidivism risk,\footnote{\url{https://www.propublica.org/article/machine-bias-risk-assessments-in-criminal-sentencing}} COMPAS \cite{larson2016we, angwin2016machine} consists of \textbf{7,217} samples with 53 attributes. We use \emph{sex} as the protected attribute to evaluate whether individuals are likely to reoffend within two years.

\section{Results}
\label{sec:results}

The performance of the BiasGuard method was evaluated across several metrics, including $EOD$, accuracy, and $DI$, and compared against a baseline and two related work methods: Threshold Optimizer and Reject Option. Table \ref{results} summarizes the results of these experiments.
\begin{table}[h]
\centering
\caption{BiasGuard results over five datasets and Protected Attribute (PA), comparing different configurations of BiasGuard to a non-mitigated baseline, Threshold Optimizer, and Reject Option post-processing methods. In 5 out of 5 experiments, BiasGuard outperformed the Equalized Odds, the leading Fairness metric, with minimal degradation in accuracy. In 3 out of 5 experiments, accuracy was improved by BiasGuard.}
\small
\begin{tabular}{|l|l|r|r|r|r|r|r|}
\hline
\textbf{\begin{tabular}[c]{@{}l@{}}Dataset\\ (PA)\end{tabular}}
& \textbf{Method}
& \textbf{\begin{tabular}[c]{@{}l@{}}Accuracy$\uparrow$\\ (mean $\pm$ std)\end{tabular}}
& \textbf{\begin{tabular}[c]{@{}l@{}}EOD$\downarrow$\\ (mean $\pm$ std)\end{tabular}}
& \textbf{$\Delta$FPR}
& \textbf{$\Delta$TPR}
& \textbf{\begin{tabular}[c]{@{}l@{}}DI$\uparrow$\\ (mean $\pm$ std)\end{tabular}}
& \textbf{Flips}
\\ \hline

\multirow{7}{*}{\begin{tabular}[c]{@{}l@{}}LAW\\ (SEX)\end{tabular}}
& BiasGuard-8
& 0.90081 $\pm$ 0.00001
& 0.02010 $\pm$ 0.00018
& 0.03578
& 0.00442
& 1.00083 $\pm$ 0.00017
& 77
\\ \cline{2-8}

& BiasGuard-6
& 0.90081 $\pm$ 0.00001
& 0.02010 $\pm$ 0.00018
& 0.03578
& 0.00442
& 1.00083 $\pm$ 0.00017
& 77
\\ \cline{2-8}

& BiasGuard-4
& 0.90081 $\pm$ 0.00001
& 0.02010 $\pm$ 0.00018
& 0.03578
& 0.00442
& 1.00083 $\pm$ 0.00017
& 77
\\ \cline{2-8}

& BiasGuard-2
& \textbf{0.90081} $\pm$ 0.00001
& \textbf{0.02010} $\pm$ 0.00018
& 0.03578
& 0.00442
& \textbf{1.00083} $\pm$ 0.00017
& \textbf{77}
\\ \cline{2-8}

& Threshold Opt
& 0.89518 $\pm$ 0.00001
& 0.02867 $\pm$ 0.00008
& 0.05143
& 0.00591
& 0.99925 $\pm$ 0.00005
& 775
\\ \cline{2-8}

& Reject Option
& 0.89999 $\pm$ 0.00001
& 0.04504 $\pm$ 0.00059
& 0.07719
& 0.01289
& 0.98460 $\pm$ 0.00049
& 265
\\ \cline{2-8}

& baseline
& 0.90062 $\pm$ 0.00000
& 0.03287 $\pm$ 0.00027
& 0.05710
& 0.00864
& 0.98633 $\pm$ 0.00023
& 0
\\ \hline

\multirow{7}{*}{\begin{tabular}[c]{@{}l@{}}SURGICAL\\ (AGE)\end{tabular}}
& BiasGuard-8
& 0.80184 $\pm$ 0.00001
& \textbf{0.03929} $\pm$ 0.00024
& 0.02039
& 0.05820
& 0.97954 $\pm$ 0.00025
& 237
\\ \cline{2-8}

& BiasGuard-6
& 0.80178 $\pm$ 0.00001
& 0.04017 $\pm$ 0.00026
& 0.02034
& 0.06000
& 0.98007 $\pm$ 0.00025
& \textbf{234}
\\ \cline{2-8}

& BiasGuard-4
& 0.80184 $\pm$ 0.00001
& 0.04000 $\pm$ 0.00025
& 0.02137
& 0.05862
& 0.97915 $\pm$ 0.00025
& 241
\\ \cline{2-8}

& BiasGuard-2
& 0.80184 $\pm$ 0.00001
& 0.03973 $\pm$ 0.00021
& 0.02060
& 0.05886
& 0.97923 $\pm$ 0.00025
& 241
\\ \cline{2-8}

& Threshold Opt
& 0.79872 $\pm$ 0.00001
& 0.08340 $\pm$ 0.00023
& 0.01532
& 0.15158
& 1.01041 $\pm$ 0.00017
& 1192
\\ \cline{2-8}

& Reject Option
& 0.80342 $\pm$ 0.00001
& 0.07134 $\pm$ 0.00008
& 0.01424
& 0.12845
& 1.01065 $\pm$ 0.00012
& 588
\\ \cline{2-8}

& baseline
& \textbf{0.80478} $\pm$ 0.00002
& 0.07222 $\pm$ 0.00009
& 0.00843
& 0.13602
& \textbf{1.01835} $\pm$ 0.00005
& 0
\\ \hline

\multirow{7}{*}{\begin{tabular}[c]{@{}l@{}}RECRUIT\\ (SEX)\end{tabular}}
& BiasGuard-8
& \textbf{0.84375} $\pm$ 0.00004
& 0.06393 $\pm$ 0.00040
& 0.02916
& 0.09870
& \textbf{1.47085} $\pm$ 0.07950
& 55
\\ \cline{2-8}

& BiasGuard-6
& 0.84275 $\pm$ 0.00005
& 0.06142 $\pm$ 0.00048
& 0.02868
& 0.09417
& 1.46556 $\pm$ 0.07974
& 55
\\ \cline{2-8}

& BiasGuard-4
& 0.84275 $\pm$ 0.00004
& \textbf{0.06082} $\pm$ 0.00046
& 0.02944
& 0.09219
& 1.46289 $\pm$ 0.07697
& 53
\\ \cline{2-8}

& BiasGuard-2
& 0.84350 $\pm$ 0.00005
& 0.06264 $\pm$ 0.00033
& 0.02992
& 0.09535
& 1.46589 $\pm$ 0.07687
& \textbf{48}
\\ \cline{2-8}

& Threshold Opt
& 0.83800 $\pm$ 0.00012
& 0.07850 $\pm$ 0.00077
& 0.05519
& 0.10130
& 0.94422 $\pm$ 0.00707
& 263
\\ \cline{2-8}

& Reject Option
& 0.83900 $\pm$ 0.00001
& 0.06763 $\pm$ 0.00076
& 0.03041
& 0.10484
& 1.04775 $\pm$ 0.14516
& 151
\\ \cline{2-8}

& baseline
& 0.84175 $\pm$ 0.00000
& 0.08136 $\pm$ 0.00040
& 0.04245
& 0.12026
& 1.11030 $\pm$ 0.20787
& 0
\\ \hline

\multirow{7}{*}{\begin{tabular}[c]{@{}l@{}}ADULT\\ (RACE)\end{tabular}}
& BiasGuard-8
& 0.86505 $\pm$ 0.00000
& 0.05086 $\pm$ 0.00011
& 0.03345
& 0.06827
& 1.42725 $\pm$ 0.71677
& 173
\\ \cline{2-8}

& BiasGuard-6
& 0.86505 $\pm$ 0.00000
& 0.05086 $\pm$ 0.00011
& 0.03345
& 0.06827
& 1.42725 $\pm$ 0.71677
& 173
\\ \cline{2-8}

& BiasGuard-4
& 0.86505 $\pm$ 0.00000
& \textbf{0.05086} $\pm$ 0.00011
& 0.03345
& 0.06827
& 1.42725 $\pm$ 0.71677
& 173
\\ \cline{2-8}

& BiasGuard-2
& 0.86505 $\pm$ 0.00000
& 0.05095 $\pm$ 0.00011
& 0.03350
& 0.06841
& \textbf{1.42756} $\pm$ 0.71749
& 173
\\ \cline{2-8}

& Threshold Opt
& 0.86122 $\pm$ 0.00002
& 0.14113 $\pm$ 0.00029
& 0.05877
& 0.22353
& 0.96249 $\pm$ 0.00439
& 801
\\ \cline{2-8}

& Reject Option
& 0.86515 $\pm$ 0.00001
& 0.07508 $\pm$ 0.00039
& 0.03238
& 0.11779
& 1.95866 $\pm$ 0.61702
& \textbf{87}
\\ \cline{2-8}

& baseline
& \textbf{0.86616} $\pm$ 0.00002
& 0.05708 $\pm$ 0.00066
& 0.03747
& 0.07670
& 1.92042 $\pm$ 0.58594
& 0
\\ \hline

\multirow{7}{*}{\begin{tabular}[c]{@{}l@{}}COMPAS\\ (SEX)\end{tabular}}
& BiasGuard-8
& 0.67951 $\pm$ 0.00020
& 0.08833 $\pm$ 0.00079
& 0.08144
& 0.09522
& 0.82583 $\pm$ 0.00103
& 238
\\ \cline{2-8}

& BiasGuard-6
& 0.68104 $\pm$ 0.00010
& 0.08903 $\pm$ 0.00096
& 0.08485
& 0.09320
& 0.82427 $\pm$ 0.00120
& 233
\\ \cline{2-8}

& BiasGuard-4
& \textbf{0.68229} $\pm$ 0.00012
& 0.08874 $\pm$ 0.00110
& 0.08442
& 0.09307
& 0.82413 $\pm$ 0.00115
& 234
\\ \cline{2-8}

& BiasGuard-2
& 0.68159 $\pm$ 0.00016
& 0.09107 $\pm$ 0.00100
& 0.08755
& 0.09458
& 0.82139 $\pm$ 0.00126
& \textbf{227}
\\ \cline{2-8}

& Threshold Opt
& 0.67480 $\pm$ 0.00008
& \textbf{0.04415} $\pm$ 0.00039
& 0.03091
& 0.05738
& \textbf{0.98486} $\pm$ 0.00391
& 375
\\ \cline{2-8}

& Reject Option
& 0.67757 $\pm$ 0.00005
& 0.13580 $\pm$ 0.00092
& 0.13351
& 0.13809
& 0.77115 $\pm$ 0.00092
& 869
\\ \cline{2-8}

& baseline
& 0.68020 $\pm$ 0.00017
& 0.13476 $\pm$ 0.00137
& 0.12941
& 0.14011
& 0.77317 $\pm$ 0.00155
& 0
\\ \hline

\end{tabular}
\label{results}
\end{table}

\subsection{Equalized Odds ($EOD$)}
BiasGuard demonstrated a superior performance in terms of $EOD$ in 5 out of the 5 experiments when compared to the baseline, in at least one of the configurations of BiasGuard. In 4 of the experiments, BiasGuard outperformed both the baseline and the related work methods. On average, BiasGuard improved $EOD$ by 31\% compared to the baseline, highlighting its effectiveness in reducing disparity in outcomes across different groups.

\subsection{Accuracy}
BiasGuard's accuracy was slightly lower than the baseline, with a small average drop of only 0.09\% across experiments. However, in 3 experiments, BiasGuard's accuracy outperformed both the baseline and the related work methods. This suggests that while BiasGuard prioritizes fairness, the trade-off in accuracy is minimal and less than related work methods.

\subsection{Disparate Impact ($DI$)}
BiasGuard also demonstrated improvements in DI in 3 experiments out of 5, when compared to both the baseline and related work methods. This suggests that BiasGuard is effective not only in enhancing fairness (EOD) but also in reducing disparities in treatment between groups, further supporting its suitability for bias mitigation.

\subsection{Comparison to Threshold Optimizer and Reject Option}
For both Threshold Optimizer and Reject Option, $EOD$ performance worsened compared to the baseline, with average decreases of 16\% and 10\%, respectively. This deterioration in fairness was accompanied by accuracy drops of 0.63\% for Threshold Optimizer and 0.21\% for Reject Option, indicating that both methods may introduce a significant trade-off between fairness and accuracy.

\begin{figure}[h]
    \centering
    \includegraphics[scale=0.33]{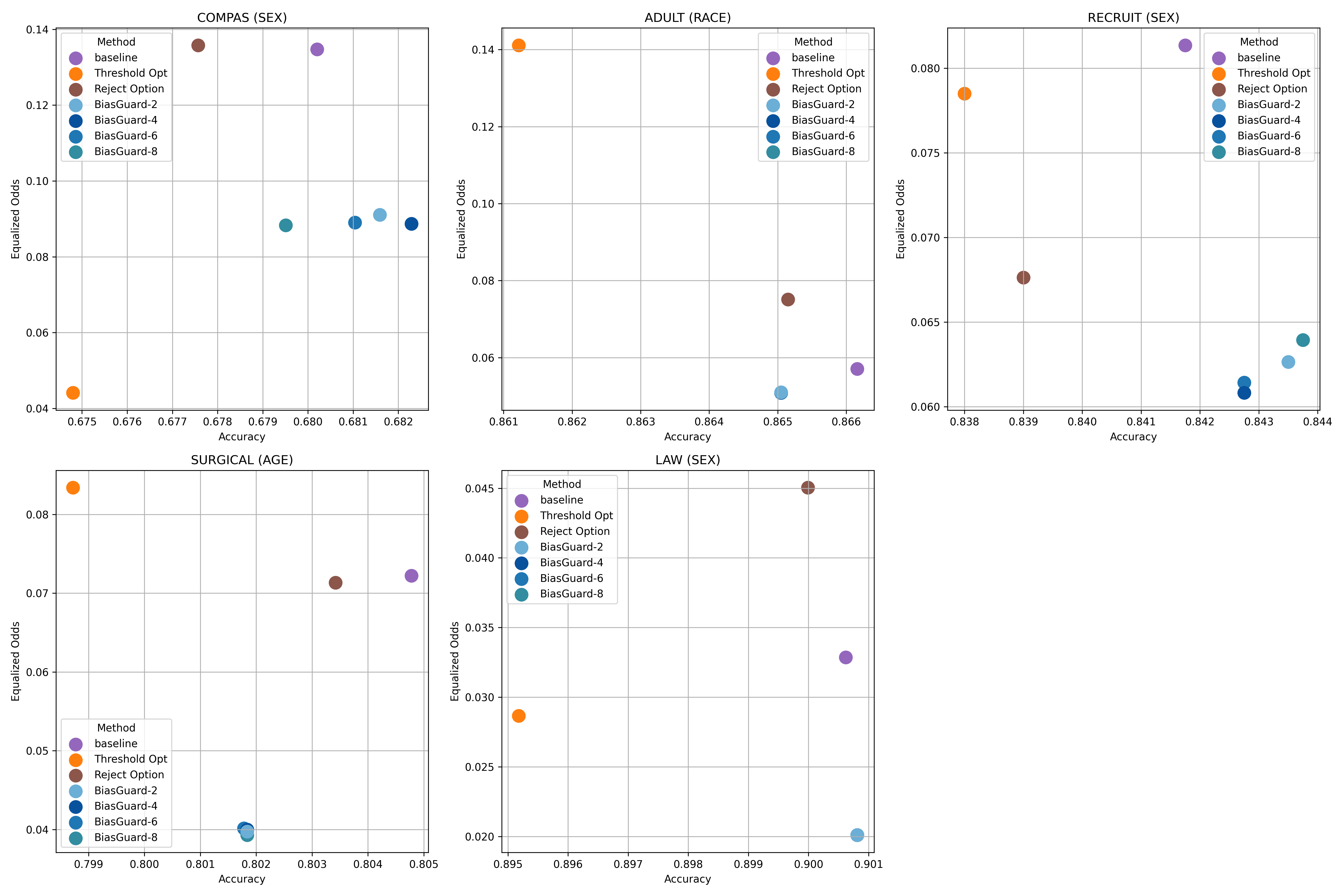}
    \caption{Tradeoff between fairness (represented by EoD) and accuracy.}
    \label{fig:tradeoff}
\end{figure}

As shown in Figure \ref{fig:tradeoff} that summarizes the results, BiasGuard outperforms the baseline in terms of fairness, with a minimal trade-off in accuracy, and demonstrates improvements in DI. 

\subsection {Sensitivity Analysis of BiasGuard Augmentation Levels}

The number of augmentations significantly influences both the performance and operational efficiency of our BiasGuard method, especially in production environments where running time is critical. We conducted a sensitivity analysis over four augmentation levels \{2, 4, 6, 8\} to understand the trade-offs between accuracy and fairness (measured via Equalized Odds). The results in Table~1 reveal that, in some datasets, different augmentation levels yield nearly identical outcomes when they involve the same ``critical flips''. In other cases, one augmentation setting achieves slightly better fairness and accuracy trade-offs, highlighting that the optimal number of augmentations varies by dataset and scenario. This choice also affects processing time, a key factor in real-time applications. Consequently, our method provides flexibility to configure the number of augmentations, enabling practitioners to choose the setting that best suits their specific constraints and objectives.

\subsection{Time Complexity Analysis}
Table \ref{tab:times} presents the running time in seconds for each of the experiments, comparing all the methods.

\begin{table}[ht!]
\centering
\caption{Inference times (seconds) for the baseline without bias mitigation, Threshold Optimizer, Reject Option, and BiasGuard methods (2, 4, 6, 8). The \#Samples column indicates each dataset’s size and the \#Attr column indicates the number of attributes used in the running time. Ratio is the average BiasGuard time divided by the baseline time.}
\small
\begin{tabular}{|l|r|r|r|r|r|c|c|c|c|r|}
\hline
\textbf{Dataset (PA)} 
& \textbf{\#Samples}
& \textbf{\#Attr}
& \textbf{baseline} 
& \textbf{Threshold Opt}
& \textbf{Reject Option}
& \multicolumn{4}{c|}{\textbf{BiasGuard}} 
& \textbf{Ratio} \\ \hline
& & & & & & \textbf{2} & \textbf{4} & \textbf{6} & \textbf{8} & \\ \hline

LAW (SEX) 
& 20798
&12
& 2       
& 3       
& 9       
& 60      
& 61      
& 62      
& 63      
& 31       
\\ \hline

SURGICAL (AGE)
& 14635
&25
& 2
& 2
& 7
& 44
& 44
& 45
& 45
& 22 
\\ \hline

RECRUIT (SEX)
& 4000
& 15
& 1
& 1
& 4
& 11
& 12
& 12
& 12
& 12 
\\ \hline

ADULT (RACE)
& 32561
& 15
& 9
& 11
& 25
& 97
& 98
& 99
& 100
& 11 
\\ \hline

COMPAS (SEX)
& 7217
& 15
& 1
& 3
& 4
& 21
& 21
& 23
& 24
& 22 
\\ \hline

\end{tabular}
\label{tab:times}
\end{table}

We compared BiasGuard’s inference time against several baselines over 5-fold evaluation and different augmentation configurations using CTGAN-based synthetic data. Our findings indicate an approximate 11–31 times increase in inference time compared to no-test-time-augmentation (TTA) or simpler baselines (Threshold Optimizer, Reject Option). However, this overhead remains practically feasible, particularly for large datasets.  
For instance, with the ADULT dataset (\(32{,}561\) test samples, which has the largest CTGAN configuration in our experiments), total inference across five folds required about 100 seconds, though it is one order of magnitude increase it is translated into a negligible per-sample overhead. Furthermore, since BiasGuard algorithm first checks whether flipping the protected attribute changes the model’s prediction, TTA-based “balancing” is only triggered if a difference is detected. Consequently, many instances do not undergo additional synthetic augmentations, and in some cases, different augmentations yield identical outcomes. In practice, the observed speed may differ from the theoretical predictions due to system constraints, engineering overhead, and implementation-specific factors
Taking into consideration that BiasGuard aims to mitigate bias for tabular datasets ML systems, the computational cost of it is acceptable for most real-world scenarios, especially given the accompanying gains in fairer decision-making.

\section{Discussion}

\subsection{Trade-Off Between Accuracy and Fairness}
In our experiments, we observed a trade-off between accuracy and fairness, a common challenge in implementing post-processing bias mitigation methods. BiasGuard achieved a 31\% improvement in fairness metrics, as measured by reductions in EOD, at the cost of a minimal 0.09\% decrease in accuracy. This trade-off reflects inherent tensions in ML models between enhancing fairness and maintaining high predictive performance, discussed thoroughly in literature ~\cite{friedler2019comparative, chen2023comprehensive}, demonstrating that trade-offs are typically present across various bias mitigation strategies, especially post-processing ones. When comparing the trade-offs to the related work methods, such as Threshold Optimizer and Reject Option, we observe a more pronounced drop in accuracy. Threshold Optimizer and Reject Option showed accuracy decreases of 0.63\% and 0.21\%, respectively. These methods not only exhibit worsened fairness but also incur larger accuracy drops compared to BiasGuard, highlighting the relatively better balance achieved by BiasGuard in maintaining predictive performance. 

For some scenarios, the trade-off observed with BiasGuard is justified by its minimal drop in accuracy, making it a more viable choice in settings where slight sacrifices in accuracy are acceptable in favor of fairness improvements.

\subsection{Flips Count}
In four out of five experiments, BiasGuard produced the fewest flips relative to other post-processing methods, yet still achieved the best EOD performance. This indicates that BiasGuard efficiently targets only those instances requiring adjustment, rather than broadly modifying predictions. Consequently, BiasGuard appears to be a more stable and reliable approach that introduces less noise while preserving most of the model’s original decisions. Its ability to minimize flips without sacrificing fairness further highlights its efficiency in bias mitigation. 

\subsection{Robustness and Agnosticism of BiasGuard}
Our experiments have demonstrated BiasGuard’s compatibility with a variety of scenarios, datasets, and protected attributes, indicating its broad applicability across different domains. Additionally, BiasGuard operates on the predictions provided by a trained classifier, rather than modifying the model’s architecture or retraining it. As a result, BiasGuard is model-agnostic: it can be seamlessly integrated with any underlying ML algorithm, from logistic regression or decision trees to deep neural networks. By leveraging only the output of the classifier, BiasGuard effectively decouples bias mitigation from model design, offering a lightweight yet powerful approach to enforcing fairness.

\subsection{Ethical Implications}
BiasGuard's ethical considerations are critical to its practical application in real-world systems. The results demonstrate consistent reductions in Equalized Odds across all groups, confirming that BiasGuard effectively addresses complex bias patterns. Additionally, its performance remained stable in datasets with imbalanced class distributions, such as the ADULT dataset (for the race protected attribute), highlighting its adaptability to dynamic real-world data scenarios.

BiasGuard's reliance on synthetic data generated by Conditional GANs (CTGAN) raises ethical considerations, particularly the risk of introducing unintended biases or over-correction. To address these concerns, we implemented rigorous monitoring of the synthetic data to ensure it accurately reflects the statistical properties of the original data, including diversity across protected and non-protected groups. This approach ensures fairness improvements for unprivileged groups without adversely affecting the outcomes for non-protected groups. Statistical analyses of prediction outcomes confirmed that BiasGuard achieves fairness without significant performance degradation or unintended disparities.

Another potential ethical challenge is over-correction, where fairness adjustments for unprivileged groups inadvertently create disadvantageous outcomes for privileged groups. BiasGuard mitigates this by balancing original predictions with those derived from synthetic augmentations rather than replacing the original predictions outright. This balance preserves the integrity of predictions for all groups while improving fairness metrics. Moreover, BiasGuard presented the least amount of flips, in comparison to other post-processing methods. Furthermore, BiasGuard’s model-agnostic nature makes it particularly suited for deployment in high-stakes applications, such as healthcare or criminal justice, where retraining the model is often infeasible. By functioning as a fairness guardrail during inference, BiasGuard enables ethical decision-making without altering the underlying model.

\section{Conclusions}
\label{sec:conclusion}
In this study, we introduced BiasGuard, an innovative method that enhances fairness in machine learning models through Test-Time Augmentation. Utilizing CTGAN to generate synthetic data with inverse protected attributes, BiasGuard addresses bias effectively without altering existing model architectures. Validated across four real-world datasets, it significantly reduces the Equalized Odds fairness metric while maintaining a reasonable accuracy trade-off.

BiasGuard acts as a dynamic fairness monitor in production environments, allowing ongoing adjustments to fairness standards in situations where model retraining is impractical. This is crucial in sectors like healthcare and finance, where decisions have significant societal impacts. By integrating synthetic instances that represent opposite protected attribute values, BiasGuard mitigates biases effectively, maintaining the model's black box integrity.

\textbf{Limitations and Future Directions:} BiasGuard introduces additional computational overhead during inference because it generates synthetic data to balance the predictions. However, this extra processing time is minimal on a per-sample basis, with inference still occurring in sub-second. Future efforts will focus on developing more efficient data generation techniques and extending the method to handle unstructured data. These advancements aim to enhance BiasGuard's scalability and versatility across diverse operational scenarios.

\section{Code Availability}
The benchmark datasets and our method's reproducible source code are available at  https://github.com/nuritci/BiasGuard 

\bibliographystyle{elsarticle-num} 
\bibliography{thesis}

\end{document}